# A Novel Approach for Shot Boundary Detection in Videos

D. S. Guru, Mahamad Suhil and P. Lolika

Department of Studies in Computer Science, Manasagangothri, Mysore,India.
dsg@compsci.uni-mysore.ac.in, mahamad45@yahoo.co.in, lolika_18@yahoo.co.in

**Abstract.** This paper presents a novel approach for video shot boundary detection. The proposed approach is based on split and merge concept. A fisher linear discriminant criterion is used to guide the process of both splitting and merging. For the purpose of capturing the between class and within class scatter we employ $2D^2$ FLD method which works on texture feature of regions in each frame of a video. Further to reduce the complexity of the process we propose to employ spectral clustering to group related regions together to a single there by achieving reduction in dimension. The proposed method is experimentally also validated on a cricket video. It is revealed that shots obtained by the proposed approach are highly cohesive and loosely coupled.

## 1  Introduction

Video shot boundary detection is a major step in the automation of content based video indexing and retrieval. A video can be viewed as a hierarchy of frames, shots and scenes as shown in Fig. 1. Shots are sequence of frames captured by a single camera in a particular time period. Identification of the transitions between two successive shots is called as shot boundary detection.

There are mainly two types of shot boundaries or shot transitions [2] namely, abrupt transitions and gradual transitions. In abrupt transitions a sudden discontinuity in the frame sequence is seen, where as in gradual transitions a slow (change) transition is seen. Fade Out, Fade In, Wipe, and dissolve are the different techniques in gradual transitions.

Various shot boundary detection algorithms are available in the literature. They are mainly classified into temporal based techniques which measure the differences in video frames with respect to time and content based techniques which detect the shot boundary by studying the content present in the frames. Many algorithms have been proposed for temporal based video segmentation including pixel based, Block based and Histogram based comparisons [6], [4], [2], [7] for both compressed and uncompressed videos; and for content based video segmentation including color [1], [8], [9], [10], texture[2], [3], [31], [11] and shape features [31], [2], [3].

Most of the shot boundary detection algorithms use different features like color histogram using local color features (LCF) [2], interest points[32] , edge based features [14], sift features [21]. Some transformations like Cosine Transform, Fourier Transform, Wavelet transforms etc., are also used [5], [20]. Block motion techniques are employed to extract the motion vectors for motion features [19], [20], [12], [22], [23]. Also, combinations of features from different modalities are used [18], [16], [13].

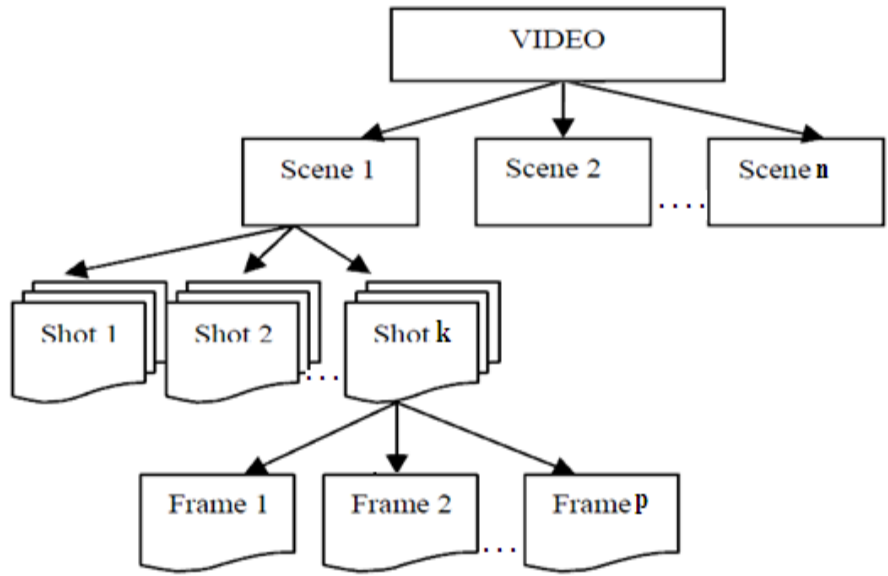

**Fig. 1.** Hierarchical structure of a video

Even though there are lots of methods available in literature for shot boundary detection, most of them have a common drawback: thresholding. So, Koumaras et al., (2005) proposed discrete cosine transform (DCT)-based and low-bit-rate encoded clips, which exploit the perceptual blockiness effect detection on each frame without using any threshold parameter. Manjunath et al., (2011) exploited the eigen gap analysis to preserve the variations among the video frames. Damnjanovic et al., in [17] explored scene change detection based on the information from eigenvectors of a similarity matrix. Goyal et al., (2009) exploited the use of split and merge mechanism to find story boundaries based on visual features and text transcripts [24].

Keeping all merits and demerits of the existing shot boundary detection algorithms in mind, in this paper, we propose a novel approach for shot boundary detection based on split and merge philosophy. The proposed method uses spectral clustering for grouping of objects in each frame subsequently apply $(2D)^2$FLD based split and merge approach for the detection of shots.

The rest of the paper is organized as follows: In section 2 the proposed split and merge based shot boundary detection is presented. The dataset and experimental results are discussed in section 3. The conclusion and future work are given in section 4.

## 2 Proposed Method

In this section we propose a novel split and merge based shot boundary detection algorithm. The main motivation is to view the video as a finite set of groups where in which each group consists of a set of adjacent frames with visually or contextually similar content preserving temporal continuity in a video. Fig. 2 shows the main steps involved in the proposed method.

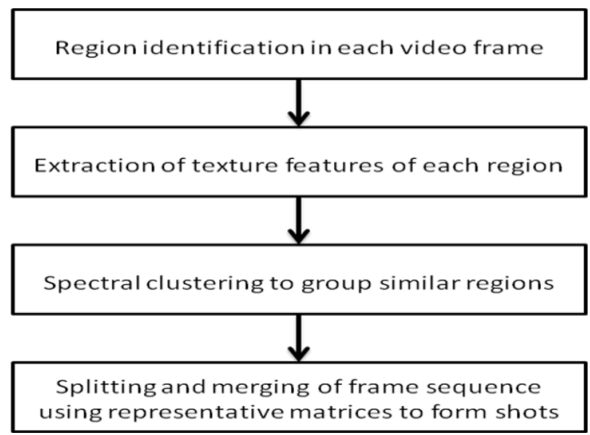

**Fig. 2.** Block diagram of the proposed method

### 2.1 Region identification in Each Video Frames

First, we identify regions in each video frame and represent them by a set of attributes to achieve dimensionality reduction. We use block division for region identification in video frames. Each video frame is first converted from RGB to a gray space and then is divided into equal sized blocks which themselves are considered as the frame regions. We resize every frame into 256 x 256 and divide it into blocks of size 32 x 32 as shown in Fig. 3.

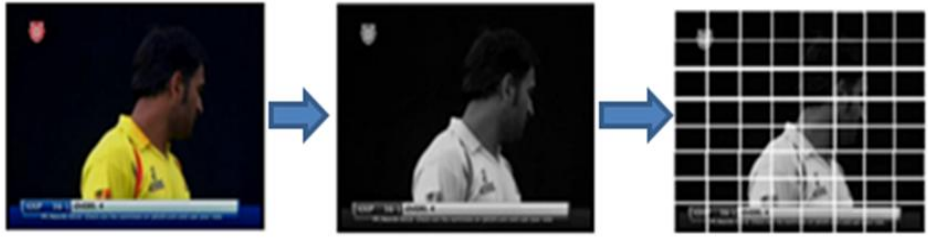

**Fig. 3.** Block based region identification

### 2.2 Textural Features for Video Frame Representation

Statistical texture features proposed by Haralick et al in [15], are used in our work for the representation of frame regions. Initially Gray Level Co-occurrence matrix (GLCM) is computed for each frame region using the pairwise occurrences of image intensities. Using the GLCM we calculate 14 different texture features for each frame region. A video frame with r regions will then be represented by r x 14 feature matrix.

Let us assume that P is the gray level co-occurrence matrix obtained from the image region r, the expressions for different texture features which we have used are as follows.

Notation
p(i,j)   (i,j)th entry in a normalized gray-tone spatial dependence matrix, = P(I,j)/R
$p_x(i)$   ith entry in the marginal-probability matrix obtained by summing the rows of p(i,j)
$N_g$   Number of distinct gray levels in the quantized image.

(1). Angular Second Moment:
$$f_i = \sum_i \sum_j \{p(i,j)\}^2$$

(2). Contrast:
$$f_2 = \sum_{n=0}^{N_g-1} n^2 \left\{ \sum_{\substack{i=1 \\ |i-j|=n}}^{N_g} \sum_{j=1}^{N_g} P(i,j) \right\}$$

(3). Correlation:
$$f_3 = \frac{\sum_i \sum_j (ij)p(i,j) - \mu_x \mu_y}{\sigma_x \sigma_y}$$

Where $\mu_x$, $\mu_y$, $\sigma_x$ and $\sigma_y$ are the means and standard deviations of $p_x$ and $p_y$.

(4). Sum of Squares: Variance

$$f_4 = \sum_i \sum_j (i - \mu)^2 p(i,j)$$

(5). Inverse Difference Moment:

$$f_5 = \sum_i \sum_j \frac{1}{1 + (i - j)^2} P(i,j)$$

(6). Sum Average:

$$f_6 = \sum_{i=2}^{2N_g} i p_{x+y}(i)$$

(7). Sum Variance:

$$f_7 = \sum_{i=1}^{2N_g} (i - f_8)^2 P_{X+Y}(j)$$

(8). Sum Entropy:

$$f_8 = - \sum_{i=2}^{2N_g} p_{x+y}(i) \log \{p_{x+y}(i)\}$$

(9). Entropy:

$$f_9 = - \sum_i \sum_j p(i,j) \log(p(i,j)).$$

(10). Difference Variance:

$$f_{10} = variance\ of\ p_{x-y}$$

(11). Difference Entropy:

$$f_{11} = - \sum_{i=0}^{N_g-1} p_{x-y}(i) \log \{p_{x-y}(i)\}$$

(12), (13). Information Measures of C

$$f_{12} = \frac{HXY - HXY1}{\max\{HX, HY\}}$$

$$f_{13} = (1 - \exp[-2.0(HXY2 - HXY)])^{1/2}$$

$$HXY = - \sum_i \sum_j p(i,j) \log(P(i,j))$$

Where HX and HY are entropies of $p_x$ and $p_y$ and ,

$$HXY1 = -\sum_i \sum_j p(i,j) \log\{p_x(i)p_y(j)\}$$

$$HXY2 = -\sum_i \sum_j p_x(i)p_y(j) \log\{p_x(i)p_y(j)\}$$

(14). Maximal Correlation Coefficient:

$$f_{14} = (second\ largest\ eigenvalue\ of\ Q)^{1/2}$$

where,

$$Q(i,j) = \sum_k \frac{p(i,k)P(j,k)}{p_x(i)p_y(k)}$$

**2.3 Clustering of Frame Regions Using Spectral Clustering**

In this step, we group the similar regions in each frame. Since spectral clustering [27],[28],[29] has been very efficiently used in the literature of image segmentation we adapt spectral clustering to group the regions in each frame using the feature matrices of each region obtained in the previous step.

The main problem in using spectral clustering is the curse of dimensionality. Because, when we try to apply spectral clustering to an image of size m x n the size of the affinity matrix will be mn x mn and when the size of an image is large it becomes a tedious job to handle such a huge affinity matrix. Our purpose is to identify different regions existing in each video frame so that we can represent the frame with some features of those regions and to make the matching of frames easier. So, instead of going for pixel based spectral clustering we go for region based spectral clustering. That is, we apply region identification to get the first impression of the regions existing in the frame and then we represent each extracted region by a set of features and spectral clustering is applied using the representative feature vectors of each region to form clusters of regions.

We are performing two levels of dimensionality reduction. First level of dimensionality reduction is achieved by identifying the regions in video frames using block division or segmentation and representing each region with a feature vector consisting of 14 texture features and the second level of dimensionality reduction is by applying spectral clustering to group the regions identified in first level. Irrespective of the size of the video frames we get a representative matrix of size k x 14 where k is the number of clusters obtained after applying spectral clustering. Hence our initial goal of classifying a sequence of frames into group of shots has become classifying the similar feature matrices into a group.

Suppose, FM(r x 14) is the feature matrix obtained after representing a video frame f with r regions where each region is described using 14 statistical texture features.

We can now say that, we have r points $x_1, \ldots, x_r$ in a $R^{14}$ dimensional space and the similarity $s_{ij} \geq 0$ between all pairs of data points $x_i$ and $x_j$ are calculated using Euclidean distance measure. We apply spectral clustering procedure proposed by Jordan et al., (2002) by treating the data as a graph G = (V, E) where data points are considered to be the set of vertices V and E is the set of edges connecting the vertices.

As a result of spectral clustering of the r points we obtain k<r clusters. We then form the representative feature matrix RM $\in R^{k \times 14}$ of the frame f, whose $i^{th}$ row is the mean of the vectors belonging to $i^{th}$ cluster.

The value of the k, the number of clusters, which is smaller than the value of r is decided based on the type of the dataset being used. It is fixed up for the entire set of frames so that the variation in the size of the representative feature matrix is maintained. We check the performances of the clustering with different values of k and an optimal k value is selected empirically.

### 2.4 Shot Boundary Detection

The proposed method detects shots in a given video using the concept of split and merge which is driven by the criterion function of Fisher's Linear Discriminant analysis. The recursive splitting and merging of the sequence of frames is done using the representative matrices obtained in the previous section. We first introduce the concept of split and merge in general and its usage in the perspective of shot boundary detection. Subsequently we move onto the concept of $(2D)^2$ FLD and how it is adapted in the proposed method for shot boundary detection.

**The Concept of Split and Merge.** Split and merge is a well known concept in image segmentation where the image is subdivided into a fixed number of regions repetitively until the predicate becomes true for all the regions. Consequently merging is done for any two adjacent regions when the predicate becomes true when they are considered as a single region. In the proposed method the same analogy is used to arrive at the shots from a given video. The larger sequence of video frames is split into 2 smaller subsequences repeatedly using the representative matrices obtained in the previous stage until the predicate becomes true. This process results in a number of smaller subsequences of the original video. Then any two adjacent subsequences are merged if the predicate is continued to be true when they are considered as a single subsequence.

**$(2d)^2$ FLD for Splitting and Merging of Video Frame Sequences.** The predicate we have considered to split the sequence of frames is the criterion function (J) of Fisher's Linear Discriminant analysis [25]. The two subsequences of frames are treated as two classes of data points and given to the FLD function; the job of FLD is to find an optimal projection axis to project the data points of two classes. If the data points are well separated in the projected space then the value of the criterion function will be maximum. But, we cannot use this criterion to split the sequence since we do not have any initial assumption about the maximum value of J such that we can fix up a threshold for it and if the value of J obtained is greater than the threshold we can allow splitting otherwise not. Actually our task is to somehow find a position for split which gives a maximum value for J. The solution for this problem is to compare the J value obtained for the sequence under consideration and its left neighboring sequence with the corresponding J value for the left sub sequence of the sequence under consideration and the same left neighboring sequence. Similarly, compare the J value obtained for the sequence under consideration and its right neighboring sequence with the corresponding J value for the right sub sequence of the sequence under consideration and the same right neighboring sequence. If both the J values in the current iteration are greater than that of previous then we allow the original sequence to get split into exactly two subsequences and update the values of J for the corresponding sequences. That is, the sequence to be checked for the possibility for splitting is passed to the FLD as two subsequences along with the two adjacent sequences, if the values of J obtained for those two subsequences with their neighboring sequences yields a greater value than the corresponding J values obtained before splitting then only that sequence will be allowed to get split.

The same procedure is considered in reverse for merging of the two adjacent subsequences. That is, any two adjacent subsequences are merged if and only if the value of J between the left subsequence and its neighboring subsequence, and the value of J between the right subsequence and its neighboring subsequence are less than or equal to the values of J at corresponding locations after merging.

The process of splitting and merging is started with a single long sequence of video frame representatives and continued until the J value gets maximized for every pair of adjacent subsequences. Each and every resultant subsequence is declared to be the shot present in the video.

Since we have the 2 dimensional representatives of each video frame, we recommend using the 2D FLD [26], [30] instead of conventional 1D FLD. Hence, We have used the most recent and efficient two dimensional Fisher's Linear Discriminant algorithm the $(2D)^2$ FLD proposed by Nagabhushan et al.,[26].

## 3  Results and Analysis

In this section, we conduct experiments on various video samples. Subsequently, quantitative evaluation of the proposed shot boundary detection method in terms of Precision, Recall and F-measures is given.

### 3.1 Dataset and Ground-Truth

**Table 1.** Dataset and Ground-truth

| Video Type | Caption of the video sequence | Frame interval | Total number of Transitions |
|---|---|---|---|
| 1. Cricket | 1. India vs. Australia 2$^{nd}$ match at Kolkata | 5400 to 5900 | 5 |
| | 2. DLF-IPL-2010 CSK Vs KXIP | 15000 to 15500 | 3 |
| 2. News | 3. NDTV–God Science and the Universe | 1000 to 1500 | 6 |
| | 4. NDTV-God Science and the Universe | 2000 to 2500 | 6 |

To test the performance of the proposed method we have considered two different types of videos, they are Cricket and News. We have considered two from each type downloaded from the internet. Manually identified shots present in each of the testing video sequence are considered as ground truth. The details of the test videos and ground truth for the evaluation of the proposed method are given in the Table 1.

### 3.2 Experimental Results

The experimental results of the proposed algorithm for the cricket and news videos are given in the Table 2. We make use of the following most popular evaluation measures to evaluate the results of the proposed method,

$$\text{Precision} = \frac{D}{(D+FA)}$$

$$\text{Recall} = \frac{D}{(D+MD)}$$

$$\text{F-measure} = \frac{2(Precision \times recall)}{Precision + Recall}$$

Where D is the number of shots correctly detected, MD is number of shots missed (MD), and FA is the number of false alarms.

**Table 2.** Experimental results of the proposed method

| TEST VIDEO SEQUENCE | Number of shots | D | MD | FA | PRESICION (P) % | RECALL (R) % | F-MEASURE F (%) |
|---|---|---|---|---|---|---|---|
| Cricket-1 | 5 | 3 | 2 | 0 | 100 | 60 | 75 |
| Cricket-2 | 3 | 3 | 0 | 0 | 100 | 100 | 100 |
| News-1 | 6 | 5 | 1 | 2 | 71.42 | 83.33 | 76.91 |
| News-2 | 6 | 4 | 2 | 3 | 57.14 | 66.66 | 61.53 |
| | | | | Average | 82.14 | 77.4975 | 78.36 |

We performed the experimentation on both cricket and news videos with different values of k, the number of clusters ranging from 2 to 64. Since, we divide each video frame into exactly 64 blocks and upon clustering at least we can expect two objects and at most there can be 64 objects. So we performed the experimentation with various values of k and we have empirically set the value of k as 6 for both types of videos. The results obtained for cricket sequences are really appreciable when compared to the results of news video. Though there are few missed detections and false alarms, the overall performance of the proposed algorithm is considerably good. Few of the missed detections and false alarms are due to the gradual transitions present in the video. The main drawback is the parameter fixation that is, fixing up of the value of k for a particular video sequence is necessary. In future we will think of an adaptive approach which can effectively solve the problem of fixing k value.

## 4 Conclusion

Shot boundary detection is the very first step in the automation of content based video indexing and retrieval. It searches for and recognizes the visual discontinuities caused by the transitions between frames, to segment a video stream into elementary uninterrupted content units for subsequent high-level semantic analysis. Despite the long research history and numerous proposed techniques, shot boundary detection is still a challenging and active area of research. In this paper we have proposed a novel approach for the shot boundary detection problem. The novelty of the proposed approach lies in the exploitation of split and merge philosophy which has been proved to be an efficient method to solve many complex problems along with two major concepts namely, fisher's linear discriminant analysis and spectral clustering. In the proposed method we have exploited the beauty of split and merge concept to split the given video sequence into smaller subsequences with temporally identical content based on another well known concept fisher's linear discriminant analysis. We have used spectral clustering to group the regions in video frames based on the textural features. Experiments are conducted on various datasets such as cricket and news. The proposed method is evaluated using Precision, Recall and F-measures and we have achieved 82.14% of Precision, 77.5% of Recall and 78.36% of F-measure for the test data set considered.